\documentclass[11pt,a4paper]{article}
\usepackage[hyperref]{acl2020}
\usepackage{times}
\usepackage{latexsym}

\usepackage{microtype}

\aclfinalcopy 
\def\aclpaperid{1960} 

\usepackage[hyperref]{acl2020}

\usepackage{amsmath, amsfonts, amssymb, amsthm, xspace, color}
\usepackage{booktabs} 

\usepackage{subfigure, multirow, tabularx} 
\usepackage{booktabs} 
\usepackage{enumitem}
\usepackage{flushend}
\usepackage[T1]{fontenc}
\usepackage{epstopdf}
\usepackage{balance}
\usepackage{graphicx}
\usepackage[noend,ruled,linesnumbered]{algorithm2e} 
\usepackage{url}
\usepackage{wrapfig,lipsum}
\usepackage{makecell}
\usepackage{wrapfig}
\usepackage{times}
\usepackage{latexsym}

\usepackage{microtype}

\newcommand{\hide}[1]{} 

\newcommand{\ie}{i.e.\xspace} 
\newcommand{\eg}{e.g.\xspace} 
\newcommand{\nop}[1]{}
\newcommand{\mquote}[1]{{``\emph{#1}''}}

\newtheorem{thm:def}{Definition}
\newtheorem{thm:eg}{Example}
\newtheorem{thm:lem}{Lemma}
\newtheorem{thm:obs}{Observation}
\newtheorem{thm:req}{Requirement}
\newtheorem{thm:prop}{Proposition}
\newtheorem{thm:principle}{Principle}
\newtheorem{thm:thm}{Theorem}
\newtheorem{thm:corollary}{Corollary}

\newcommand{\abs}[1]{\mathopen| #1 \mathclose|}			

\def \P {\mathbf{P}}

\def \D {\mathcal{D}}

\newcommand{\Our}{\mbox{CGExpan}\xspace}
\newcommand{\OurNoCN}{\mbox{CGExpan-NoCN}\xspace}
\newcommand{\OurNoFilter}{\mbox{CGExpan-NoFilter}\xspace}

\newcommand{\OurMRR}{\mbox{CGExpan-MRR}\xspace}
\newcommand{\OurComb}{\mbox{CGExpan-Comb}\xspace}

\definecolor{midnightgreen}{rgb}{0.0, 0.29, 0.33}
\definecolor{orange}{RGB}{255,127,0}

\usepackage{bbm}
\usepackage{adjustbox}
\usepackage[super]{nth}

\title{Empower Entity Set Expansion via Language Model Probing}

\author{Yunyi Zhang$^{1}$, Jiaming Shen$^{1}$, Jingbo Shang$^{2}$, Jiawei Han$^1$\\
$^1$ University of Illinois at Urbana-Champaign, IL, USA \\
$^2$ University of California San Diego, CA, USA \\
$^{1}$\{\texttt{yzhan238}, \texttt{js2}, \texttt{hanj}\}\texttt{@illinois.edu} $~~~$ $^{2}$ \texttt{jshang@ucsd.edu} \\
}

\date{}

\begin{document}
\maketitle
\begin{abstract}

Entity set expansion, aiming at expanding a small seed entity set with new entities belonging to the same semantic class, is a critical task that benefits many downstream NLP and IR applications, such as question answering, query understanding, and taxonomy construction. 
Existing set expansion methods bootstrap the seed entity set by adaptively selecting context features and extracting new entities. 
A key challenge for entity set expansion is to avoid selecting ambiguous context features which will shift the class semantics and lead to accumulative errors in later iterations. 
In this study, we propose a novel iterative set expansion framework that leverages automatically generated class names to address the semantic drift issue.
In each iteration, we select one positive and several negative class names by probing a pre-trained language model, and further score each candidate entity based on selected class names. 
Experiments on two datasets show that our framework generates high-quality class names and outperforms previous state-of-the-art methods significantly.

\end{abstract}

\section{Introduction}\label{sec:intro}

Entity set expansion aims to expand a small set of seed entities (\eg, \{\mquote{United States},  \mquote{China},  \mquote{Canada}\}) with new entities (\eg, \mquote{United Kingdom}, \mquote{Australia}) belonging to the same semantic class (\ie, \texttt{Country}). 
The entities so discovered may benefit a variety of NLP and IR applications, such as question answering~\cite{Wang2008AutomaticSE},  query understanding~\cite{Hua2017UndersandST}, taxonomy construction~\cite{Shen2018HiExpanTT}, and semantic search~\cite{Xiong2017ExplicitSR, Shen2018EntitySS}. 

Most existing entity set expansion methods bootstrap the initial seed set by iteratively selecting context features (\eg, co-occurrence words~\cite{Pantel2009WebScaleDS}, unary patterns~\cite{Rong2016egoset}, and coordinational patterns~\cite{Mamou2018TermSE}), while extracting and ranking new entities.
A key challenge to set expansion is to avoid selecting \emph{ambiguous patterns} that may introduce erroneous entities from other non-target semantic classes.
Take the above class \texttt{Country} as an example, we may find some ambiguous patterns like \mquote{* located at} (which will match more general \texttt{Location} entities) and \mquote{match against *} (which may be associated with entities in the \texttt{Sports Club} class). 
Furthermore, as bootstrapping is an iterative process, those erroneous entities added at early iterations may shift the class semantics, leading to inferior expansion quality at later iterations. 
Addressing such ``semantic drift'' issue without requiring additional user inputs (\eg, mutually exclusive classes~\cite{Curran2007MinimisingSD} and negative example entities~\cite{Jindal2011LearningFN}) remains an open research problem. 

 \begin{figure}[!t]
   \centering
   \centerline{\includegraphics[width=0.48\textwidth]{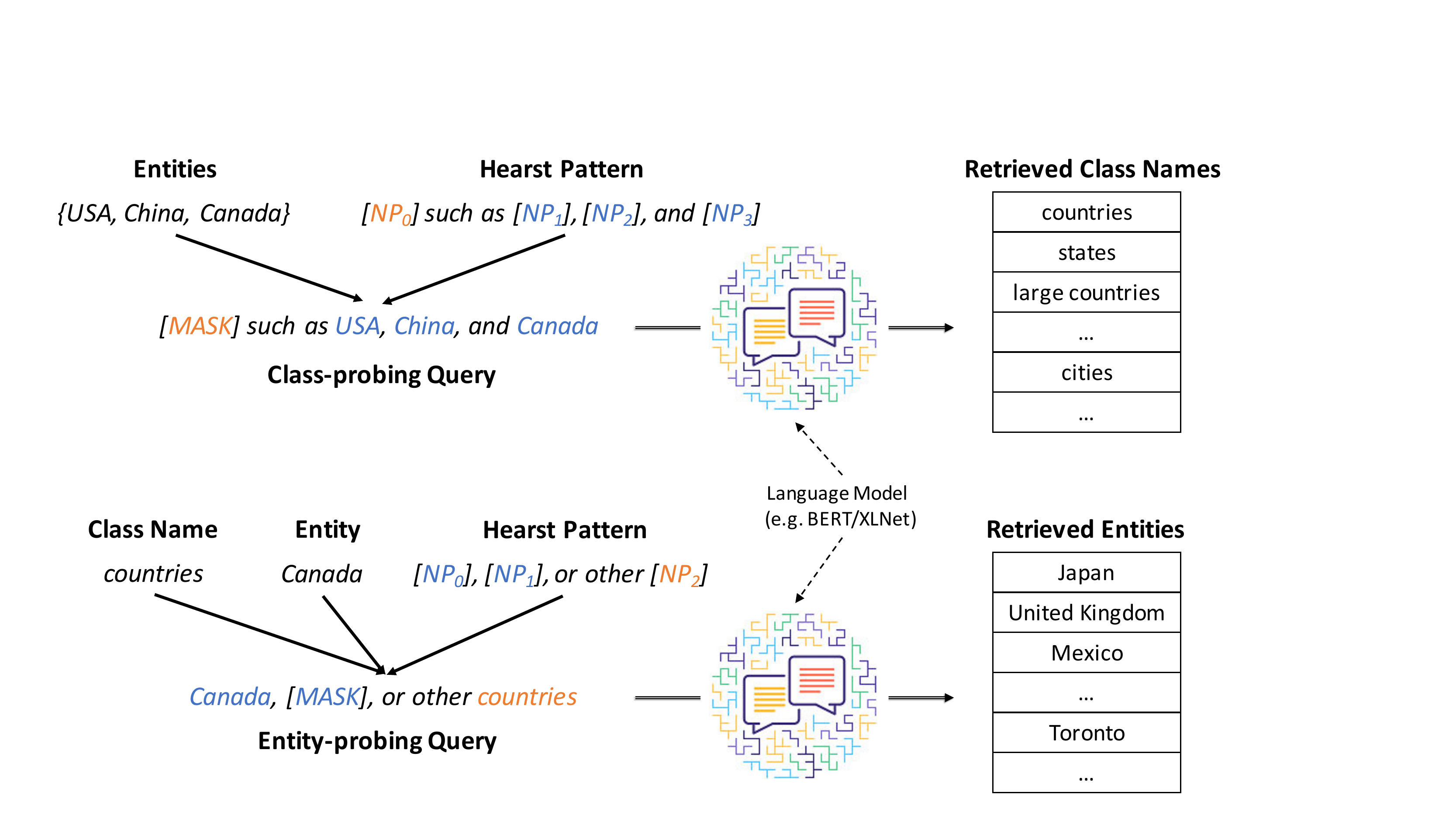}}
   \vspace{-0.2cm}
   \caption{Examples of class-probing and entity-probing queries generated based on Hearst patterns.}
   \label{fig:intro}
   \vspace{-0.3cm}
 \end{figure}

In this study, we propose to empower entity set expansion with class names automatically generated from pre-trained language models~\cite{Peters2018DeepCW,Devlin2019BERTPO, Yang2019XLNetGA}. 
Intuitively, knowing the class name is ``country'', instead of ``state'' or ``city'', can help us identify \emph{unambiguous patterns} and eliminate erroneous entities like \mquote{Europe} and \mquote{New York}. 
Moreover, we can acquire such knowledge (\ie, positive and negative class names) by probing a pre-trained language model automatically without relying on human annotated data. 

Motivated by the above intuition, we propose a new iterative framework for entity set expansion that consists of three modules:
(1) The first, \emph{class name generation} module, constructs and submits class-probing queries (\eg, ``[\textsc{Mask}] such as USA, China, and Canada.'' in Fig.~\ref{fig:intro}) to a language model for retrieving a set of candidate class names. 
(2) The second, \emph{class name ranking} module, builds an entity-probing query for each candidate class name and retrieves a set of entities. The similarity between this retrieved set and the current entity set serves as a proxy for the class name quality, based on which we rank all candidate class names.
An unsupervised ensemble technique~\cite{Shen2017SetExpanCS} is further used to improve the quality of final ranked list from which we select one best class name and several negative class names.
(3) The third, \emph{class-guided entity selection} module, scores each entity conditioned on the above selected class names and adds top-ranked entities into the currently expanded set. 
As better class names may emerge in later iterations, we score and rank all entities (including those already in the expanded set) at each iteration, which helps alleviate the semantic drift issue. 

\smallskip
\noindent \textbf{Contributions.}~
In summary, this study makes the following contributions: 
(1) We propose a new set expansion framework that leverages class names to guide the expansion process and enables filtration of the entire set in each iteration to resolve the semantic drift issue;  
(2) we design an automatic class name generation algorithm that outputs high-quality class names by dynamically probing pre-trained language models; and
(3) experiments on two public datasets from different domains demonstrate the superior performance of our approach compared with state-of-the-art methods.

\section{Background}\label{sec:problem}
\vspace{-0.1cm}

In this section, we provide background on language models and define the entity set expansion problem.

    \subsection{Language Model}
    
    A standard language model (LM) inputs a word sequence $\mathbf{w} = [w_1, w_2, \dots, w_n]$ and assigns a probability $\P(\mathbf{w})$ to the whole sequence. 
    Recent studies~\cite{Peters2018DeepCW,Devlin2019BERTPO, Yang2019XLNetGA} found that language models, simply trained for next word or missing word prediction, can generate high quality \emph{contextualized} word representations which benefit many downstream applications. 
    Specifically, these language models will output an embedding vector for each \emph{word appearance} in a specific context that is usually the entire sentence where the target word occurs, rather than just words appearing before the target word. 
    Therefore, we can also view a LM as a model that inputs a word sequence $\mathbf{w}$ and outputs a probability $\P(w_i)=\P(w_i|w_1,  \dots, w_{i-1}, w_{i+1}, \dots, w_n)$ to any position $1 \leq i \leq n$. 
    Currently, \citet{Devlin2019BERTPO} propose BERT and train the language model with two objectives: (1) a cloze-filling objective which randomly substitutes some words with a special [\textsc{Mask}] token in the input sentence and forces LM to recover masked words, and (2) a binary classification objective that guides LM to predict whether one sentence directly follows another (sentence). 
    BERT leverages Transformer \cite{Vaswani2017AttensionIA} architecture and is learned on English Wikipedia as well as BookCorpus.
    More LM architectures are described in Section~\ref{sec:related_work}. 
    
    \subsection{Problem Formulation}
    We first define some key concepts and then present our problem formulation.
    
    \noindent \textbf{Entity.} An entity is a word or a phrase that refers to a real-world instance. For example, \mquote{U.S.} refers to the country: United States. 
    
    \noindent \textbf{Class Name.} A class name is a text representation of a semantic class. For instance, \texttt{country} could be a class name for the semantic class that includes entities like \mquote{United States} and \mquote{China}.
    
    \noindent \textbf{Probing Query.} A probing query is a word sequence containing one [\textsc{Mask}] token. 
    In this work, we utilize Hearst patterns~\cite{hearst1992automatic} to construct two types of probing queries: (1) A \emph{class-probing} query aims to predict the class name of some given entities (e.g., ``[\textsc{Mask}] such as United States and China''), and (2) an \emph{entity-probing} query aims to retrieve entities that fit into the mask token (e.g., ``countries such as [\textsc{Mask}] and Japan''). 
    
    \noindent \textbf{Problem Formulation.} Given a text corpus $\D$ and a seed set of user-provided entities, we aim to 
    output a ranked list of entities that belong to the same semantic class. 
    
    \noindent \textbf{Example 1.} Given a seed set of three countries $\{$\mquote{United States}, \mquote{China}, \mquote{Canada}$\}$, we aim to return a ranked list of entities belonging to the same \texttt{country} class such as \mquote{United Kingdom}, \mquote{Japan}, and \mquote{Mexico}.
\section{Class-Guided Entity Set Expansion}\label{sec:method}

We introduce our class-guided entity set expansion framework in this section.
First, we present our class name generation and ranking modules in Sections~\ref{subsec:generation} and \ref{subsec:ranking}, respectively.
Then, we discuss how to leverage class names to guide the iterative expansion process in Section~\ref{subsec:expansion}.

\begin{figure*}[!t]
    \centering
    \centerline{\includegraphics[width=1.0\textwidth]{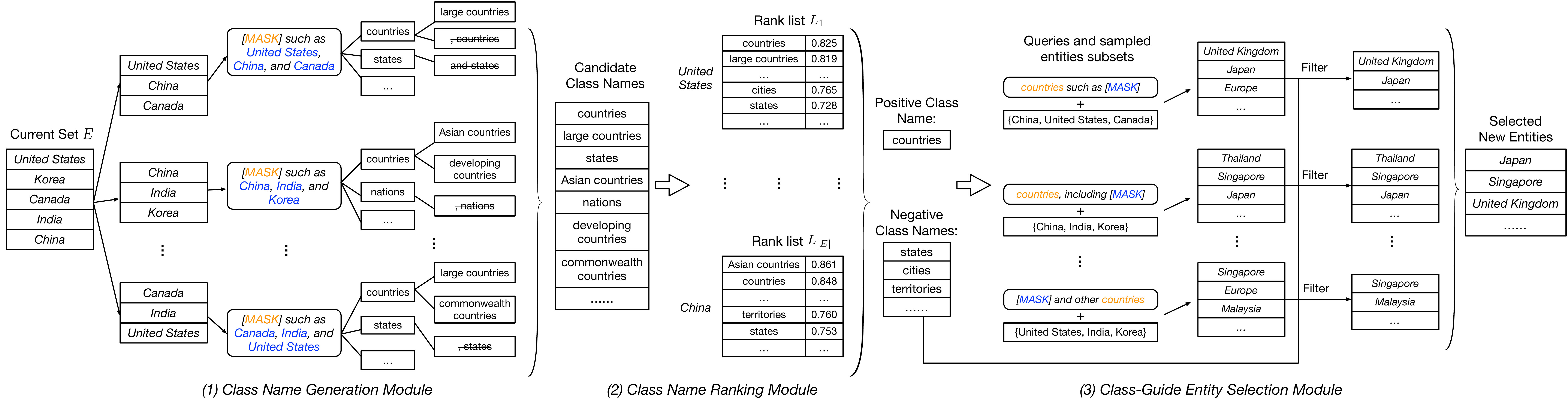}}
    \vspace{-0.2cm}
    \caption{Overview of \emph{one iteration} in \Our framework.}
    \label{fig:framework}
    \vspace{-0.3cm}
\end{figure*}

\subsection{Class Name Generation}\label{subsec:generation}

    The class name generation module inputs a small collection of entities and generates a set of candidate class names for these entities. 
    We build this module by automatically constructing class-probing queries and iteratively querying a pre-trained LM to obtain multi-gram class names. 
    
    First, we notice that the class name generation goal is similar to the hypernymy detection task which aims to find a general hypernym (\eg, ``mammal'') for a given specific hyponym (\eg, ``panda''). 
    Therefore, we leverage the six Hearst patterns~\cite{hearst1992automatic}\footnote{\footnotesize For example, the pattern ``$NP_y$ such as $NP_a$'' indicates that noun phrase $y$ is a hypernym of noun phrase $a$.}, widely used for hypernymy detection, to construct the class-probing query. 
    More specifically, we randomly select three entities in the current set as well as one Hearst pattern (out of six choices) to construct one query. 
    For example, we may choose entities \{\mquote{China}, \mquote{India}, \mquote{Japan}\} and pattern ``$NP_y$ such as $NP_a$, $NP_b$, and $NP_c$'' to construct the query ``[\textsc{Mask}] such as China, India, and Japan''. 
    By repeating such a random selection process, we can construct a set of queries and feed them into pre-trained language models to obtain predicted masked tokens which are viewed as possible class names. 
    
    The above procedure has one limitation---it can only generate unigram class names. 
    To obtain multi-gram class names, we design a modified beam search algorithm to iteratively query a pre-trained LM. 
    Specifically, after we query a LM for the first time and retrieve top $K$ most likely words (for the masked token), we construct $K$ new queries by adding each retrieved word after the masked token. 
    Taking the former query  ``[\textsc{Mask}] such as China, India, and Japan'' as an example, we may first obtain words like ``countries'', ``nations'', and then construct a new query ``[\textsc{Mask}] \underline{countries} such as China, India, and Japan''. 
    Probing the LM again with this new query, we can get words like ``Asian'' or ``large'', and obtain more fine-grained class names like ``Asian countries'' or ``large countries''. 
    We repeat this process for maximum three times and keep all generated class names that are noun phrases\footnote{\footnotesize Therefore, class names likes ``and countries'' and ``, countries'' are filtered out.}.
    As a result, for each Hearst pattern and randomly selected three entities from the current set, we will obtain a set of candidate class names.
    Finally, we use the union of all these sets as our candidate class name pool, denoted as $\mathcal{C}$.  
    Note that in this module, we focus on the recall of candidate class name pool $\mathcal{C}$, without considering its precision, since the next module will further rank and select these class names based on the provided text corpus.

\subsection{Class Name Ranking}\label{subsec:ranking}

	In this module, we rank the above generated candidate class names to select one best class name that represents the whole entity set and some negative class names used in the next module to filter out wrong entities. 
	A simple strategy is to rank these class names based on the number of times it has been generated in the previous module.
	However, such a strategy is sub-optimal because short unigram class names always appear more frequently than longer multi-gram class names.
	Therefore, we propose a new method below to measure how well each candidate class name represents the entity set. 
    
    	First, we introduce a corpus-based similarity measure between an entity $e$ and a class name $c$.
	Given the class name $c$, we first construct 6 \emph{entity-probing} queries by masking the hyponym term in six Hearst patterns\footnote{\footnotesize For example, a query for class name ``countries'' is ``\underline{countries} such as [\textsc{Mask}]''.}, and query a pre-trained LM to obtain the set of six [\textsc{Mask}] token embeddings, denoted as $X_c$. 
	Moreover, we use $X_e$ to denote the set of all contextualized representations of the entity $e$ in the given corpus. 
	Then, we define the similarity between $e$ and $c$, as:
	\begin{equation}\label{eq:softmatch}
	\small
        M^k(e,c) = \frac{1}{k}\max_{X\subseteq X_e, \abs{X}=k}\sum_{\mathbf{x}\in X}\max_{\mathbf{x}'\in X_c} \mathbf{cos}(\mathbf{x}, \mathbf{x}'), 
    \end{equation}
    where $\mathbf{cos}(\mathbf{x}, \mathbf{x}')$ is the cosine similarity between two vectors $\mathbf{x}$ and $\mathbf{x}'$. 
    The inner $\max$ operator finds the maximum similarity between each occurrence of $e$ and the set of entity-probing queries constructed based on $c$. 
    The outer $\max$ operator identifies the top-$k$ most similar occurrences of $e$ with the queries and then we take their average as the final similarity between the entity $e$ and the class name $c$. 
    This measure is analogous to finding $k$ best \emph{occurrences} of entity $e$ that matches to any of the probing queries of class~$c$, and therefore it improves the previous similarity measures that utilize only the \emph{context-free} representations of entities and class names (\eg, Word2Vec). 
     
    After we define the entity-class similarity score, we can choose one entity in the current set and obtain a ranked list of candidate class names based on their similarities with this chosen entity. 
    Then, given an entity set $E$, we can obtain $|E|$ ranked lists, $L_{1}, L_{2}, \dots, L_{|E|}$, one for each entity in $E$.
    Finally, we follow~\cite{Shen2017SetExpanCS} and aggregate all these lists to a final ranked list of class names based on the score $s(c) = \sum_{i=1}^{|E|} \frac{1}{r_{c}^{i}}$, where $r_{c}^{i}$ indicates the rank position of class name $c$ in ranked list $L_{i}$. 
    This final ranked list shows the order of how well each class name can represent the current entity set.
    Therefore, we choose the best one that ranks in the first position as the positive class , denoted as $c_{p}$. 
    
    Aside from choosing the positive class name $c_{p}$, we also select a set of negative class names for the target semantic class to help bound its semantics. 
    To achieve this goal, we assume that entities in the initial user-provided seed set $E^{0}$ definitely belong to the target class.
    Then, we choose those class names that rank lower than $c_{p}$ in \emph{all} lists corresponding to entities in $E^{0}$, namely \{$L_{i} | e_{i} \in E^{0}$\}, and treat them as the negative class names. 
    We refer to this negative set of class names as $C_N$ and use them to guide the set expansion process below.

\subsection{Class-Guided Entity Selection}\label{subsec:expansion}

    In this module, we leverage the above selected positive and negative class names to help select new entities to add to the set. 
    We first introduce two entity scoring functions and then present a new rank ensemble algorithm for entity selection. 
    
    The first function utilizes the positive class name $c_p$ and calculates each entity $e_{i}$'s score : 
    \begin{equation}\label{eq:sim1}
    	\small
        \text{score}_{i}^{loc} = M^k(e_i, c_{p}),
    \end{equation}
    where $M^{k}$ is defined in Eq.~(\ref{eq:softmatch}). 
    We refer to this score as a \emph{\underline{loc}al} score because it only looks at top-$k$ best occurrences in the corpus where the contextualized representation of entity $e_i$ is most similar to the representation of class name $c_{q}$. 
    
    The second scoring function calculates the similarity between each candidate entity and existing entities in the current set, based on their \emph{context-free} representations. 
    For each entity $e$, we use the average of all its contextualized embedding vectors as its context-free representation, denoted as $\mathbf{v}_{e}$. 
    Given the current entity set $E$, we first sample several entities from $E$, denoted as $E_{s}$, and calculate the score for each candidate entity $e_{i}$ as:
    \begin{equation}\label{eq:sim2}
    	\small
        \text{score}_{i}^{glb} = \frac{1}{\abs{E_s}}\sum_{e\in E_s} \textbf{cos}(\mathbf{v}_{e_i}, \mathbf{v}_{e}).
    \end{equation}
    Note here we sample a small set $E_{s}$ (typically of size 3), rather than using the entire set $E$. Since the current entity set $E$ may contain wrong entities introduced in previous steps, we do not use all the entities in $E$ and compute the candidate entity score only once. Instead, we randomly select multiple subsets of entities from the current set $E$, namely $E_s$,  obtain a ranked list of candidate entities for each sampled subset, and aggregate all ranked lists to select the final entities. Such a sampling strategy can reduce the effect of using wrong entities in $E$, as they are unlikely to be sampled multiple times, and thus can alleviate potential errors that are introduced in previous iterations.
    We refer to this score as a \emph{\underline{gl}o\underline{b}al} score because it utilizes context-free representations which better reflect entities' overall positions in the embedding space and measure the entity-entity similarity in a more global sense. 
    Such a global score complements the above local score and we use their geometric mean to finally rank all candidate entities: 
    \begin{equation}\label{eq:sim}
    	\small
        \text{score}_{i} = \sqrt{ \text{score}_{i}^{loc} \times  \text{score}_{i}^{glb}}.
    \end{equation}

    As the expansion process iterates, wrong entities may be included in the set and cause semantic drifting. 
    We develop a novel rank ensemble algorithm that leverages those selected class names to improve the quality and robustness of entity selection.
    First, we repeatedly sample $E_{s}$ (used for calculating $\text{score}_{i}^{glb}$ in Eq.~(\ref{eq:sim2})) $T$ times from current entity set $E$, and obtain $T$ entity ranked lists $\{R^m\}_{m=1}^{T}$.
    Second, we follow the class name ranking procedure in Section~\ref{subsec:ranking} to obtain $|E|$ class ranked lists $\{L^n\}_{n=1}^{|E|}$, one for each entity $e_{i} \in E$. 
    Note here each $L^n$ is actually a ranked list over $\{c_p\} \cup C_N$, namely the set of selected one positive class name and all negative class names. 
    Intuitively, an entity belonging to our target semantic class should satisfy two criteria: (1) it appears at the top positions in multiple entity ranked lists, and (2) within its corresponding class ranked list, the selected best class name $c_p$ should be ranked above any one of the negative class name in $C_{N}$. 
    Combining these two criteria, we define a new rank aggregation score as follows:
    \begin{align}\label{eq:mmrr}
    	\small
        S(e_i) &= \sum_{t=1}^T \left(\mathbbm{1}(e_i \in E)+s^t(e_i)\right)\nonumber \\ 
        & \times \mathbbm{1}(r_{c_p}^i<\min_{c'\in C_N}r_{c'}^i),
    \end{align}
    where $\mathbbm{1}(\cdot)$ is an indicator function, $r_c^i$ is the rank of class name $c$ in entity $e_i$'s ranked list $L_c^i$, and $s^t(e_i)$ the individual aggregation score of $e_i$ deduced from the ranked list $R^t$, for which we test two aggregation methods: (1) mean reciprocal rank, where \begin{equation}
    	\small
        s^t(e_i) = \frac{1}{r_i^t}
    \end{equation}
    and $r_i^t$ is the rank of entity $e_i$ in the $t$-th ranked list $R^t$; and (2) the combination of scores (CombSUM), where  
    \begin{equation}
    	\small
        s^t(e_i) = \frac{\text{score}^t_i - \min_{e_j\in R^t}{\text{score}^t_j}}{\max_{e_j\in R^t}{\text{score}^t_j} - \min_{e_j\in R^t}{\text{score}^t_j}}
    \end{equation}
    is the ranking score of $e_i$ in the ranked list $R^t$ after min-max feature scaling.
    
    To interpret Eq.~\ref{eq:mmrr}, the first summation term reflects our criterion (1) and its inner indicator function ensuring an entity in the current set $E$ prone to have a large rank aggregation score if not been filtered out below. 
    The second term reflects our criterion (2) by using an indicator function that filters out all entities which are more similar to a negative class name than the positive class name. 
    Note here we calculate the aggregation score for all entities in the vocabulary list, including those already in the current set $E$, and it is possible that some entity in $E$ will be filtered out because it has 0 value in the second term. 
    This makes a huge difference comparing with previous iterative set expansion algorithms which all assume that once an entity is included in the set, it will stay in the set forever. 
    Consequently, our method is more robust to the semantic drifting issue than previous studies.

    \smallskip
    \noindent \textbf{Summary.}
    Starting with a small seed entity set, we iteratively apply the above three modules to obtain an entity ranked list and add top-ranked entities into the set. We repeat the whole process until either (1) the expanded set reaches a pre-defined target size or (2) the size of the set does not increase for three consecutive iterations. Notice that, by setting a large target size, more true entities belonging to the target semantic class will be selected to expand the set, which increases the recall, but wrong entities are also more likely to be included, which decreases the precision. However, as the output of the set expansion framework is a ranked list, the most confident high-quality entities will still be ranked high in the list.

\section{Experiments}\label{sec:exp}
    
    \begin{table}[t]
    \centering
    \scalebox{0.9}{
        \small
        \begin{tabular}{ccccc}
            \toprule
            \textbf{Dataset} & \textbf{\# Test Queries} & \textbf{\# Entities} & \textbf{\# Sentences} \\
            \midrule
            Wiki & 40 & 33K & 1.50M \\
            \midrule
            APR & 15 & 76K & 1.01M \\
            \bottomrule
        \end{tabular}
    }
    \vspace{-0.2cm}
    \caption{Datasets statistics}\label{table:dateset}
    \vspace{-0.3cm}
    \end{table}

    \subsection{Experiment Setup}
    
    \noindent \textbf{Datasets.} We conduct our experiments on two public benchmark datasets widely used in previous studies~\cite{Shen2017SetExpanCS,Yan2019LearningTB}: 
    (1) \textbf{Wiki}, which is a subset of English Wikipedia articles\nop{from May 2011}, and (2) \textbf{APR}, which contains all news articles published by Associated Press and Reuters in 2015. 
    Following the previous work, we adopt a phrase mining tool, AutoPhrase \cite{Shang2018AutomatedPM}, to construct the entity vocabulary list from the corpus, and select the same 8 semantic classes for the Wiki dataset as well as 3 semantic classes for the APR dataset. 
    Each semantic class has 5 seed sets and each seed set contains 3 entities.
    Table~\ref{table:dateset} summarizes the statistics for these datasets.
    
    \begin{table*}
    \centering
    \scalebox{0.85}{
    \begin{tabular}{lcccccc}
    \toprule
    \multirow{2}{*}{\textbf{Methods}}   & \multicolumn{3}{c}{\textbf{Wiki}}                         & \multicolumn{3}{c}{\textbf{APR}}                          \\
                               & MAP@10         & MAP@20         & MAP@50         & MAP@10         & MAP@20         & MAP@50         \\ 
                               
    \midrule
    Egoset~\cite{Rong2016egoset}                   & 0.904          & 0.877          & 0.745          & 0.758          & 0.710          & 0.570          \\ 
    SetExpan~\cite{Shen2017SetExpanCS}                   & 0.944            & 0.921          & 0.720          & 0.789          & 0.763          & 0.639          \\ 
    SetExpander~\cite{Mamou2018TermSE}                & 0.499          & 0.439          & 0.321          & 0.287          & 0.208          & 0.120          \\
    CaSE~\cite{Yu2019CorpusbasedSE}                       & 0.897          & 0.806          & 0.588          & 0.619          & 0.494          & 0.330          \\
    MCTS~\cite{Yan2019LearningTB} & 0.980{\small$^{\nabla}$} & 0.930{\small$^{\nabla}$} & 0.790{\small$^{\nabla}$} & 0.960{\small$^{\nabla}$} & 0.900{\small$^{\nabla}$} & 0.810{\small$^{\nabla}$}  \\
    \midrule
    \OurNoCN & 0.968          & 0.945          & 0.859          & 0.909          & 0.902          & 0.787          \\
    \OurNoFilter & 0.990          & 0.975          & 0.890          & 0.979          & 0.962          & 0.892          \\
    \OurComb & 0.991          & 0.974          & 0.895          & 0.983          & 0.984          & 0.937          \\
    \OurMRR       & \textbf{0.995} & \textbf{0.978} & \textbf{0.902} & \textbf{0.992} & \textbf{0.990} & \textbf{0.955} \\ \bottomrule
    \end{tabular}
    }
    \vspace*{-0.2cm}
    \caption{\label{table:results} Mean Average Precision on Wiki and APR. ``$^{\nabla}$'' means the number is directly from the original paper.}
    \vspace*{-0.3cm}
    \end{table*}

    \noindent \textbf{Compared methods.} We compare the following corpus-based entity set expansion methods. 
    \begin{enumerate}[leftmargin=*,nosep]
        \item Egoset~\cite{Rong2016egoset}: This is a multifaceted set expansion system using context features and Word2Vec embeddings. The original framework aims to expand the set in multiple facets.   Here we treat all expanded entities as in one semantic class due to little ambiguity in the seed set.
        \item SetExpan~\cite{Shen2017SetExpanCS}: This method iteratively selects skip-gram context features from the corpus and develops a rank ensemble mechanism to score and select entities.
        \item SetExpander~\cite{Mamou2018TermSE}: This method trains different embeddings based on different types of context features and leverages additional human-annotated sets to build a classifier on top of learned embeddings to predict whether an entity belongs to the set.
        \item CaSE~\cite{Yu2019CorpusbasedSE}: This method combines entity skip-gram context feature and embedding features to score and rank entities once from the corpus. 
        The original paper has three variants and we use the CaSE-W2V variant since it is the best model claimed in the paper. 
        \item MCTS~\cite{Yan2019LearningTB}: This method bootstraps the initial seed set by combing the Monte Carlo Tree Search algorithm with a deep similarity network to estimate delayed feedback for pattern evaluation and to score entities given selected patterns. 
        \item \Our: This method is our proposed \underline{C}lass-\underline{G}uided Set \underline{Expan}sion framework, using BERT~\cite{Devlin2019BERTPO} as the pre-trained language model. We include two versions of our full model, namely \OurComb and \OurMRR, that use the combination of score and mean reciprocal rank for rank aggregation, respectively.
        \item \OurNoCN: An ablation of \Our that excludes the class name guidance. Therefore, it only incorporates the average BERT representation to select entities. 
        \item \OurNoFilter: An ablation of \Our that excludes the negative class name selection step and uses only the single positive class name in the entity selection module. 
    \end{enumerate}
    
    \begin{table}
    \centering
    \scalebox{0.88}{
        \small
        \begin{tabular}{lccc}
            \toprule
            \textbf{\Our vs. Other}   & \textbf{MAP@10}     & \textbf{MAP@20}    & \textbf{MAP@50}      \\
            \midrule
            vs. SetExpan         & 100\%     & 94.5\%     & 87.3\%        \\ 
            vs. \OurNoFilter     & 100\%      & 94.5\%      & 58.2\%        \\ 
            vs. \OurNoCN     & 100\%     & 94.5\%      & 70.9\%       \\ 
            \bottomrule
        \end{tabular}
    }
    \vspace{-0.2cm}
    \caption{\label{table:compare} Ratio of seed entity set queries on which the first method reaches better or the same performance as the second method.}
    \vspace{-0.3cm}
    \end{table}
    
    \smallskip
    \noindent \textbf{Evaluation Metric.} 
    We follow previous studies and evaluate set expansion results using Mean Average Precision at different top $K$ positions (MAP@$K$) as below:
    \begin{displaymath}
    \small
    \text{MAP@}\textit{K} = \frac{1}{|Q|} \sum_{q \in Q} \text{AP}_{K} (L_{q}, S_{q}),
    \end{displaymath}
    \normalsize
    where $Q$ is the set of all seed queries and for each query $q$, we use $\text{AP}_{K} (L_{q}, S_{q})$ to denote the traditional average precision at position $K$ given a ranked list of entities $L_{q}$ and a ground-truth set $S_{q}$. 
    
    \smallskip
    \noindent \textbf{Implementation Details.} 
    For CGExpan, we use BERT-base-uncased\footnote{\footnotesize In principle, other masked LMs such as RoBERTa and XLNet can also be used in our framework.} as our pre-trained LM.
    For parameter setting, in the class name generation module (Sec.~\ref{subsec:generation}), we take top-3 predicted tokens in each level of beam search and set the maximum length of generated class names up to 3. 
    When calculating the similarity between an entity and a class name (Eq.~\ref{eq:softmatch}), we choose $k=5$, and will later provide a parameter study on $k$ in the experiment. 
    Also, since MAP@$K$ for $K=10, 20, 50$ are typically used for set expansion evaluations, we follow the convention and choose 50 as the target set size in our experiments.\footnote{\footnotesize The code and data are available at \url{https://github.com/yzhan238/CGExpan}}
    
    \begin{table}[t]
    \centering
    \scalebox{0.88}{
        \small
        \begin{tabular}{lcc}
        \toprule
        \multirow{2}{*}{\textbf{Methods}}   & \textbf{Wiki}                         & \textbf{APR}                          \\
                                   & MAP@\{10/20/50\}         &  MAP@\{10/20/50\} \\ 
                                   
        \midrule
        Oracle-Full   & 0.991/0.976/0.891          & 1.000/1.000/0.964          \\
        Oracle-NoFilter & 0.994/0.983/0.887          & 0.988/0.966/0.894          \\
        \Our       & 0.995/0.978/0.902 & 0.992/0.990/0.955 \\ 
        \bottomrule
        \end{tabular}
    }
        \vspace{-0.2cm}
        \caption{\label{table:compare_oracle} Compared to oracle models knowing ground truth class names, \Our automatically generates class names and achieves comparative performances.}
        \vspace{-0.3cm}
    \end{table}
    
    \subsection{Experiment Results}
    
    \noindent \textbf{Overall Performance.} 
    Table \ref{table:results} shows the overall performance of different entity set expansion methods. 
    We can see that \Our along with its ablations in general outperform all the baselines by a large margin.
    Comparing with SetExpan, the full model \Our achieves 24\% improvement in MAP@50 on the Wiki dataset and 49\% improvement in MAP@50 on the APR dataset, which verifies that our class-guided model can refine the expansion process and reduce the effect of erroneous entities on later iterations. 
    In addition, \OurNoCN outperforms most baseline models, meaning that the pre-trained LM itself is powerful to capture entity similarities.
    However, it still cannot beat \OurNoFilter model, which shows that we can properly guide the set expansion process by incorporating generated class names. 
    Moreover, by comparing our full model with \OurNoFilter, we can see that negative class names indeed help the expansion process by estimating a clear boundary for the target class and filtering out erroneous entities. 
    Such an improvement is particularly obvious on the APR dataset. 
    The two versions of our full model overall have comparable performance, but \OurMRR consistently outperforms \OurComb. To explain such a difference, empirically we observe that high-quality entities tend to rank high in most of the ranked lists. Therefore, we use the MRR version for the rest of our experiment, denoted as \Our.

    \begin{table*}
    \centering
    \scalebox{0.85}{
    \small
    \begin{tabular}{|c|c|c|c|}
    \hline
    Seed Entity Set                                                & Ground True Class Name  & Positive Class Name & Negative Class Names           \\ \hline
    \{\mquote{Intel}, \mquote{Microsoft}, \mquote{Dell}\}                               & \texttt{company}          & \texttt{company}             & \texttt{product}, \texttt{system}, \texttt{bank}, ...     \\ \hline
    \{\mquote{United States}, \mquote{China}, \mquote{Canada}\}                         & \texttt{country}          & \texttt{country}             & \texttt{state}, \texttt{territory}, \texttt{island}, ...  \\ \hline
    \{\mquote{ESPNews}, \mquote{ESPN Classic}, \mquote{ABC}\}                           & \texttt{tv channel}       & \texttt{television network}  & \texttt{program}, \texttt{sport}, \texttt{show}, ...      \\ \hline
    \{\mquote{NHL}, \mquote{NFL}, \mquote{American league}\}                           & \texttt{sports league}    & \texttt{professional league} & \texttt{sport}, \texttt{competition}, ...        \\ \hline
    \{\mquote{democratic}, \mquote{labor}, \mquote{tories}\}                            & \texttt{party}            & \texttt{political party}     & \texttt{organization}, \texttt{candidate}, ...   \\ \hline
    \{\mquote{Hebei}, \mquote{Shandong}, \mquote{Shanxi}\}                              & \texttt{Chinese province} & \texttt{chinese province}    & \texttt{city}, \texttt{country}, \texttt{state}, ...      \\ \hline
    \begin{tabular}[c]{@{}c@{}}\{\mquote{tuberculossi}, \mquote{Parkinson's disease},\\ \mquote{esophageal cancer}\}\end{tabular}
                        & \texttt{disease}          & \texttt{chronic disease}     & \texttt{symptom}, \texttt{condition}, ... \\ \hline
    \{\mquote{Illinois}, \mquote{Arizona}, \mquote{California}\}                        & \texttt{US state}         & \texttt{state}               & \texttt{county}, \texttt{country}, ...           \\ \hline
    \end{tabular}
    }
    \vspace{-0.2cm}
    \caption{\label{table:class} Class names generated for seed entity sets. The \nth{2} column is the ground true class name in the original dataset. The \nth{3} and \nth{4} columns are positive and negative class names predicted by \Our, respectively.}
    \vspace{-0.3cm}
    \end{table*}

    \smallskip
    \noindent \textbf{Fine-grained Performance Analysis.} 
    Table~\ref{table:compare} reports more fine-grained comparison results between two methods. Specifically, we calculate the ratio of seed entity set queries (out of total 55 queries) on which one method achieves better or the same performance as the other method. 
    We can see that \Our clearly outperforms SetExpan and its two variants on the majority of queries. 
    In Table~\ref{table:compare_oracle}, we further compare \Our with two ``oracle'' models that have the access to ground truth class names. Results show that \Our can achieve comparative results as those oracle models, which indicates the high quality of generated class names and effectiveness of \Our. 
    
\begin{figure}[!t]
       \centering
       \centerline{\includegraphics[width=0.48\textwidth]{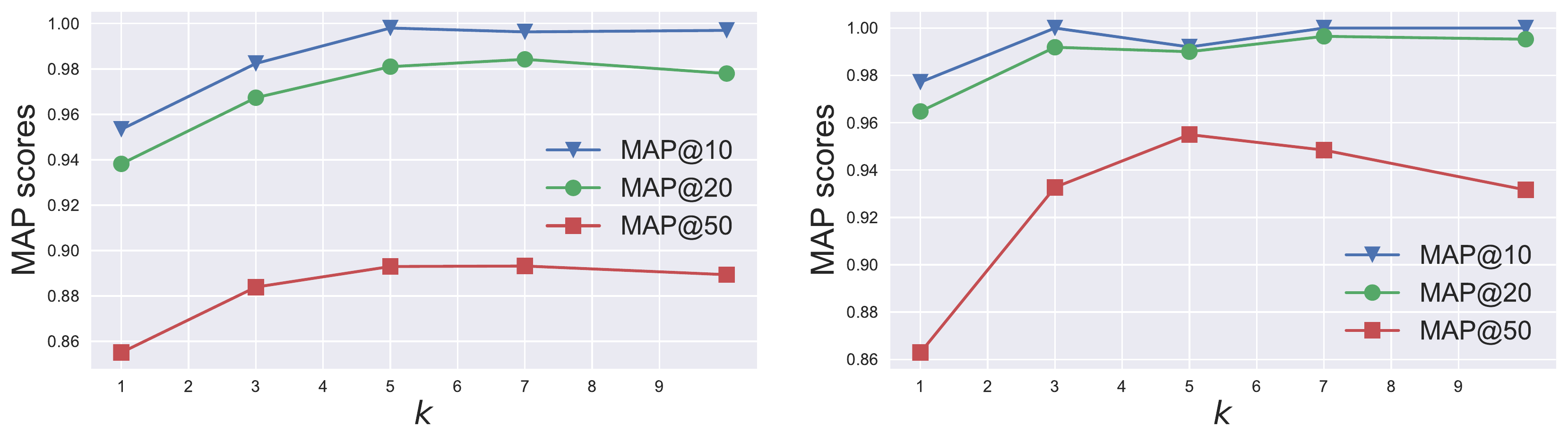}}
       \vspace{-0.2cm}
       \caption{Performance for different $k$ values on Wiki (left) and APR (right).}
       \label{fig:param_study}
       \vspace{-0.3cm}
     \end{figure}

    \smallskip
    \noindent \textbf{Parameter Study.} 
    In \Our, we calculate the similarity between an entity and a class name based on its $k$ occurrences that are most similar to the class name (cf. Eq.~(\ref{eq:softmatch})). 
    Figure~\ref{fig:param_study} studies how this parameter $k$ would affect the overall performance.
    We find that the model performance first increases when $k$ increases from 1 to 5 and then becomes stable (in terms of MAP@10 and MAP@20) when $k$ further increases to 10. 
    Overall, we find $k=5$ is enough for calculating entity-class similarity and \Our is insensitive to $k$ as long as its value is larger than 5. 
    
    \subsection{Case Studies}
    
    \noindent \textbf{Class Name Selection.} 
    Table~\ref{table:class} shows some results of our class name ranking module for several queries from different semantic classes in the Wiki dataset. 
    We see that \Our is able to select the correct class name and thus injects the correct semantics in later entity selection module. 
    Moreover, as shown in the last column, \Our can identify several negative class names that provide a tight boundary for the target semantic class, including \texttt{sports} and \texttt{competition} for \texttt{sport league} class, as well as \texttt{city} and \texttt{country} for \texttt{Chinese province} class. 
    These negative class names help \Our avoid adding those related but erroneous entities into the set. 
    
    From Table~\ref{table:class} we can see that it happens when the predicted positive class name is not exactly the ground true class name in the original dataset. However, since we use both the generated class names and currently expanded entities as guidance and select new entities according to the context features in the provided corpus, those imperfect class names can still guide the set expansion process and perform well empirically. 
    
    Also, in principle, synonyms of the positive class name can be wrongly selected as negative class names, which also happens but very rarely in our experiments. However, since these synonyms consistently rank lower than the positive one for the initial seeds based on the given corpus, they are indeed not good class names for this specific corpus. Thus, misclassifying them will not have much influence on the performance of our model. 
    
    \smallskip
    \noindent \textbf{Entity Selection.}
    Table~\ref{table:cases} shows expanded entity sets for two sample queries. 
    After correctly predicting true positive class names and selecting relevant negative class names, \Our utilizes them to filter out those related but erroneous entities, including two TV shows in \texttt{television network} class and three entities in \texttt{political party} class. 
    As a result, \Our can outperform \OurNoFilter. 
    
    \begin{table*}
    \centering
    \scalebox{0.8}{
    \begin{tabular}{|c|c|c|c|c|c|c|}
    \hline
    Seed Entity Set                                                                                                    & \multicolumn{2}{c|}{\Our}                           & \multicolumn{2}{c|}{\OurNoCN}                       & \multicolumn{2}{c|}{\OurNoFilter}        \\ \hline
                                                                                                                       & 1  & \mquote{Pb}                                                & 1  & \mquote{NBC}                                                & 1  & \mquote{Pb}                                      \\ \cline{2-7} 
                                                                                                                       & 2  & \mquote{ABC}                                               & 2  & \mquote{CBS}                                                & 2  & \mquote{Mtv}                                     \\ \cline{2-7} 
                                                                                                                       & 3  & \mquote{CBS}                                               & 3  & \mquote{Disney Channel}                                     & 3  & \mquote{ABC}                                     \\ \cline{2-7} 
                                                                                                                       &    & ...                                               &    & ...                                                &    & ...                                     \\ \cline{2-7} 
                                                                                                                       & 35 & \mquote{Telemundo}                                         & 35 & {\color[HTML]{FE0000} \mquote{ESPN Radio}*}                  & 35 & \mquote{MyNetworkTV}                             \\ \cline{2-7} 
                                                                                                                       & 36 & \mquote{Fox Sports Net}                                    & 36 & \mquote{BBC America}                                        & 36 & \mquote{ESPN2}                                   \\ \cline{2-7} 
                                                                                                                       & 37 & \mquote{Dateline NBC}                                      & 37 & \mquote{G4}                                                 & 37 & {\color[HTML]{FE0000} \mquote{the Today Show}*}   \\ \cline{2-7} 
                                                                                                                       & 38 & \mquote{Channel 4}                                         & 38 & {\color[HTML]{FE0000} \mquote{Sirius Satellite Radio}*}      & 38 & {\color[HTML]{FE0000} \mquote{Access Hollywood}*} \\ \cline{2-7} 
                                                                                                                       & 39 & \mquote{The History Channel}                               & 39 & \mquote{TNT}                                                & 39 & \mquote{Cartoon Network}                         \\ \cline{2-7} 
    \multirow{-10}{*}{\begin{tabular}[c]{@{}c@{}}\{\mquote{ESPN}, \\ \mquote{Discovery Channel}, \\ \mquote{Comedy Central}\}\end{tabular}}       &    & ...                                               &    & ...                                                &    & ...                                     \\ \hline
                                                                                                                       & 1  & \mquote{republican}                                        & 1  & \mquote{national party}                                     & 1  & \mquote{republican}                              \\ \cline{2-7} 
                                                                                                                       & 2  & \mquote{likud}                                             & 2  & \mquote{labour party}                                       & 2  & \mquote{likud}                                   \\ \cline{2-7} 
                                                                                                                       & 3  & \mquote{liberal democrats}          & 3  & {\color[HTML]{FE0000} \mquote{gop establishment}*}           & 3  & \mquote{liberal democrats}                       \\ \cline{2-7} 
                                                                                                                       &    & ...                                               &    & ...                                                &    & ...                                     \\ \cline{2-7} 
                                                                                                                       & 40 & \mquote{komeito}                    & 40 & {\color[HTML]{FE0000} \mquote{republican jewish coalition}*} & 40 & {\color[HTML]{FE0000} \mquote{young voters}*}     \\ \cline{2-7} 
                                                                                                                       & 41 & \mquote{centrist liberal democrats} & 41 & {\color[HTML]{FE0000} \mquote{british parliament}*}          & 41 & \mquote{bjp}                                     \\ \cline{2-7} 
                                                                                                                       & 42 & {\color[HTML]{FE0000} \mquote{aipac}*}                      & 42 & {\color[HTML]{FE0000} \mquote{tea party patriots}*}          & 42 & {\color[HTML]{FE0000} \mquote{religious}*}        \\ \cline{2-7} 
                                                                                                                       & 43 & \mquote{aam aadmi party}                                   & 43 & \mquote{centrist liberal democrats}                         & 43 & {\color[HTML]{FE0000} \mquote{congress}*}         \\ \cline{2-7} 
                                                                                                                       & 44 & \mquote{ennahda}                    & 44 & {\color[HTML]{FE0000} \mquote{federal government}*}          & 44 & \mquote{lib dem}                                 \\ \cline{2-7} 
    \multirow{-10}{*}{\begin{tabular}[c]{@{}c@{}}\{\mquote{democratic party},\\ \mquote{republican party},\\ \mquote{labor party}\}\end{tabular}} &    & ...                                               &    & ...                                                &    &                                         \\ \hline
    \end{tabular}
    }
    \vspace{-0.2cm}
    \caption{\label{table:cases} Expanded entity sets for two sample queries, with erroneous entities colored {\color[HTML]{FE0000} red} and marked with a ``*''.}
    \vspace{-0.3cm}
    \end{table*}

\section{Related Work}\label{sec:related_work}

\noindent \textbf{Entity Set Expansion.} 
Traditional entity set expansion systems such as Google Sets~\cite{tong2008system} and SEAL ~\cite{Wang2007LanguageIndependentSE, Wang2008IterativeSE} typically submit a query consisting of seed entities to a general-domain search engine and extract new entities from retrieved web pages. 
These methods require an external search engine for online seed-oriented data collection, which can be costly. Therefore, more recent studies propose to expand the seed set by offline processing a corpus. 
These corpus-based set expansion methods can be categorized into two general approaches: (1) \emph{one-time entity ranking} which calculates entity distributional similarities and ranks all entities once without back and forth refinement~\cite{Mamou2018TermSE,Yu2019CorpusbasedSE}, and (2) \emph{iterative bootstrapping} which aims to bootstrap the seed entity set by iteratively selecting context features and ranking new entities~\cite{Rong2016egoset, Shen2017SetExpanCS,Yan2019LearningTB, Zhu2019FUSEMS, Huang2020GuidingCS}. 
Our method in general belongs to the later category.
Finally, there are some studies that incorporate extra knowledge to expand the entity set, including negative examples~\cite{Curran2007MinimisingSD, McIntosh2008WeightedME, Jindal2011LearningFN}, semi-structured web table~\cite{Wang2015ConceptEU}, and external knowledge base~\cite{Yu2019CourseCE}. 
Particularly, \citet{Wang2015ConceptEU} also propose to use a class name to help expand the target set.
However, their method requires a \emph{user-provided} class name and utilizes web tables as additional knowledge, while our method can automatically generate both positive and negative class names and utilize them to guide the set expansion process. 

\smallskip
\noindent \textbf{Language Model Probing.}
Traditional language models aim at assigning a probability for an input word sequence. 
Recent studies have shown that by training on next word or missing word prediction task, language models are able to generate contextualized word representations that benefit many downstream applications. 
ELMo~\cite{Peters2018DeepCW} proposes to learn a BiLSTM model that captures both forward and backward contexts.
BERT~\cite{Devlin2019BERTPO} leverages the Transformer architecture and learns to predict randomly masked tokens in the input word sequence and to classify the neighboring relation between pair of input sentences. 
Based on BERT's philosophy, RoBERTa~\cite{liu2019roberta} conducts more careful hyper-parameter tuning to improve the performance on downstream tasks. 
XLNet~\cite{Yang2019XLNetGA} further combines the ideas from ELMo and BERT and develops an autoregressive model that learns contextualized representation by maximizing the expected likelihood over permutations of the input sequence. 

Aside from generating contextualized representations, pre-trained language models can also serve as knowledge bases when being queried appropriately. 
\citet{petroni2019language} introduce the language model analysis probe and \emph{manually define} probing queries for each relation type.
By submitting those probing queries to a pre-trained LM, they show that we can retrieve relational knowledge and achieve competitive performance on various NLP tasks. 
More recently, \citet{bouraoui2019inducing} further analyze BERT's ability to store relational knowledge by using BERT to automatically select high-quality templates from text corpus for new relation prediction. 
Comparing with previous work, in this paper, we show that probing pre-trained language model works for entity set expansion task, and we propose a new entity set expansion framework that combines corpus-independent LM probing with corpus-specific context information for better expansion performance.

\section{Conclusions}\label{sec:conclusion}
In this paper, we propose a new entity set expansion framework that can use a pre-trained LM to generate candidate class names for the seed set, rank them according to the provided text corpus, and guide the entity selection process with the selected class names.
Extensive experiments on the Wiki and APR datasets demonstrate the effectiveness of our framework on both class name prediction and entity set expansion. 
In the future, we plan to expand the method scope from expanding concrete entity sets to more abstract concept sets.
For example, we may expand the set \{\mquote{machine translation}, \mquote{information extraction}, \mquote{syntactic parsing}\} to acquire more \texttt{NLP task} concepts. 
Another interesting direction is to generate a class name hierarchy via language model probing. 

\section*{Acknowledgments}
Research was sponsored in part by US DARPA KAIROS Program No. FA8750-19-2-1004 and SocialSim Program No.  W911NF-17-C-0099, National Science Foundation IIS 16-18481, IIS 17-04532, and IIS-17-41317, and DTRA HDTRA11810026. Any opinions, findings, and conclusions or recommendations expressed herein are those of the authors and should not be interpreted as necessarily representing the views, either expressed or implied, of DARPA or the U.S. Government. The U.S. Government is authorized to reproduce and distribute reprints for government purposes notwithstanding any copyright annotation hereon. We thank anonymous reviewers for valuable feedback.

\bibliography{main}
\bibliographystyle{acl_natbib}

\newpage
\appendix

\section{Six Hearst Patterns} \label{sec:hearst}
In our framework, we use the following six Hearst patterns originally proposed in \cite{hearst1992automatic}:
\begin{enumerate}
    \item $NP_y$ such as $NP_a$
    \item such $NP_y$ as $NP_a$
    \item $NP_a$ or other $NP_y$
    \item $NP_a$ and other $NP_y$
    \item $NP_y$, including $NP_a$
    \item $NP_y$, especially $NP_a$
\end{enumerate}
Within each pattern, the noun phrase $y$ is a hypernym of the noun phrase $a$.

\section{Implementation Details of CGExpan} \label{sec:inplementation}
In our experiments, we use the pre-trained BERT-base-uncased model provided in Huggingface's \texttt{Transformers} library~\cite{Wolf2019HuggingFacesTS} and do not perform any further fine-tuning on our datasets. 
To get the contextual embedding vector for an entity in the corpus, we first substitute the entity with the [\textsc{Mask}] token and use the output of the last layer for the [\textsc{Mask}] token as the embedding vector for this entity in this context. 
By doing this, we can better compare an entity's context with our probing queries (e.g. ``countries such as [\textsc{Mask}].'')

In the class name generation module, each time we will randomly sample three entities in the current set and one Hearst pattern to construct the initial probing query for the root node. 
When we grow the tree, we keep top three predicted tokens by LM at each node and we only generate class names up to tri-grams. 
Furthermore, in each iteration, we sample 30 entities subsets from current set and take the union of the generated class names for these 30 class name trees as our candidate class names. When calculating similarity between an entity and a class name, we use $k=5$ occurrences of the entity that are most similar to any entity-probing query constructed with the class name in most experiment, and we have put a parameter study on the effect of various $k$ values on the performance of our method. 
Finally, in the class-guided entity selection module, we randomly sample 18 size-3 entity subsets to get 18 entity rank lists for later rank ensemble.

In our experiments, we use the Wiki and APR datasets that are published by SetExpan~\cite{Shen2017SetExpanCS}\footnote{\scriptsize \url{https://github.com/mickeystroller/SetExpan}} and following previous work to process the corpus.

\section{Implementation Details of Baselines} \label{sec:inplementation}

We use the open-source implementations of baseline methods SetExpan~\cite{Shen2017SetExpanCS}, SetExpander~\cite{Mamou2018TermSE}\footnote{\scriptsize \url{https://github.com/NervanaSystems/nlp-architect/tree/master/nlp_architect/solutions/set_expansion}}, and CaSE~\cite{Yu2019CorpusbasedSE}\footnote{\scriptsize \url{https://github.com/PxYu/entity-expansion}}.

\end{document}